\documentclass[lettersize,journal]{IEEEtran}
\usepackage{amsmath,amsfonts}
\usepackage{algorithmic}
\usepackage{algorithm}
\usepackage{array}
\usepackage[caption=false,font=normalsize,labelfont=sf,textfont=sf]{subfig}
\usepackage{textcomp}
\usepackage{stfloats}
\usepackage{url}
\usepackage{verbatim}
\usepackage{graphicx}
\usepackage{cite}
\usepackage{multirow}
\usepackage{booktabs}
\begin{document}

\title{Improving Long-tailed Object Detection with Image-Level Supervision by Multi-Task Collaborative Learning}

\author{Bo~Li$^{\dagger}$,
        Yongqiang~Yao$^{\dagger}$,
        Jingru~Tan$^*$,
        Xin~Lu,
        Fengwei~Yu,
        Ye~Luo,
        and Jianwei Lu
\thanks{Bo Li and Ye Luo are with Tongji University, Shanghai, China. E-mail: 1911030@tongji.edu.cn, yeluo@tongji.edu.cn.}
\thanks{Jingru Tan is with Shanghai Jiao Tong University, Shanghai, China. E-mail: tanjingru120@gmail.com.}
\thanks{Yongqiang Yao, Xin Lu, and Fengwei Yu are with SenseTime Research, Shanghai, China. E-mail: soundbupt@gmail.com, luxin@sensetime.com, yufengwei@sensetime.com}
\thanks{Jianwei Lu is with Shanghai University of Traditional Chinese Medicine, Shanghai, China. E-mail: jwlu33@shutcm.edu.cn.}
\thanks{$^*$ Corresponding author. $^{\dagger}$ Equal Contribution.}
}



\maketitle

\begin{abstract}
Data in real-world object detection often exhibits the long-tailed distribution.
Existing solutions tackle this problem by mitigating the competition between the head and tail categories.
However, due to the scarcity of training samples, tail categories are still unable to learn discriminative representations.
Bringing more data into the training may alleviate the problem, but collecting instance-level annotations is an excruciating task. 
In contrast, image-level annotations are easily accessible but not fully exploited.
In this paper, we propose a novel framework CLIS  (multi-task Collaborative Learning with Image-level Supervision), which leverage image-level supervision to enhance the detection ability in a multi-task collaborative way. 
Specifically, there are an object detection task (consisting of an instance-classification task and a localization task) and an image-classification task in our framework, responsible for utilizing the two types of supervision.
Different tasks are trained collaboratively by three key designs:
(1) task-specialized sub-networks that learn specific representations of different tasks without feature entanglement.
(2) a siamese sub-network for the image-classification task that shares its knowledge with the instance-classification task, resulting in feature enrichment of detectors.
(3) a contrastive learning regularization that maintains representation consistency, bridging feature gaps of different supervision. 
Extensive experiments are conducted on the challenging LVIS dataset. Without sophisticated loss engineering, CLIS achieves an overall AP of 31.1 with 10.1 point improvement on tail categories, establishing a new state-of-the-art. Code will be at \url{https://github.com/waveboo/CLIS}.

\end{abstract}

\begin{IEEEkeywords}
Long-tailed Object Detection, Multi-Task Collaborative Learning, Image-Level Supervision
\end{IEEEkeywords}

\section{Introduction}
\label{sec:intro}

\IEEEPARstart{G}{eneral} object detection ~\cite{faster_rcnn2015ren,focalloss2017lin} has achieved great progress thanks to deep neural networks. However, these methods are mainly performed on balanced datasets(\emph{e.g.,} PASCAL VOC~\cite{pascalvoc2010everingham} and MS COCO~\cite{coco2014lin}), in which the instance numbers of all categories are close.
\begin{figure}[ht]
  \begin{center}
  \includegraphics[width=0.98\linewidth]{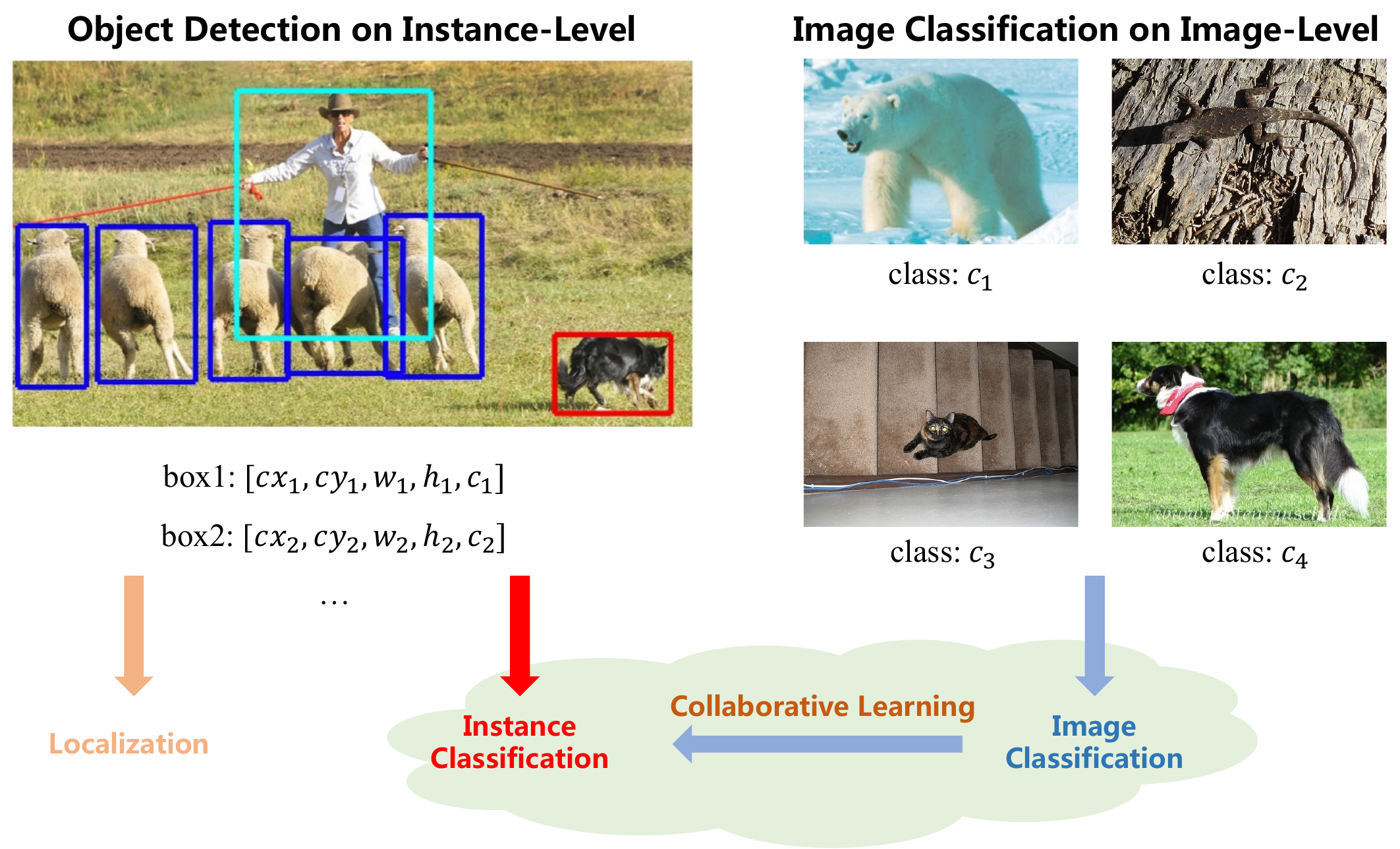}
  \end{center}
  \caption{An overview of our framework. Our framework involves three tasks, namely the localization task, the instance-classification task, and the image-classification task. Two classification tasks are learned collaboratively for improving the detection ability.}
  \label{fig:overlook}
\end{figure}
When it comes to a more realistic scenario (\emph{e.g.,} LVIS~\cite{lvis2019gupta}), the categories usually follow a long-tailed distribution, where a few head categories contain plenty of instances while most tail categories are instance-scarce. In practice, tail categories often show poor performance~\cite{lvis2019gupta,openimages2018kuznetsova}. The main difficulty lies in two aspects: On the one hand, tail categories are easily overwhelmed by the dominant head categories due to extreme imbalance. On the other hand, deep learning methods are data-hungry, while the number of instances for tail categories may not be sufficient to learn good feature representations.

Most existing solutions try to address the long-tailed problem from the first perspective. They re-balance the contribution of different categories by data re-sampling~\cite{lvis2019gupta,chang2021_sample_object_level}, cost-sensitive learning~\cite{eql2020tan,eqlv22021tan,efl2022li,seesawloss2021wang}, decoupled training~\cite{simcal2020wang,bags2020li}, and so on. 
However, all these methods investigate the long-tailed problem under limited bounding-box annotations. The performance improvement mainly comes from a seesaw game that decreases the score ranks of the head categories and increases them for tail categories~\cite{dave2021devilap}. Tail categories are still unable to learn discriminative feature representations. If we train detectors with only limited tail category annotations, generalization and performance can not be promised.



Different from the rebalance-based methods, we hope to solve the long-tailed problem from the second perspective, exploiting more training data to alleviate the instance-scarce problem for better feature representations.
However, collecting images with instance-level supervision(\emph{i.e.} bounding-box annotations) is a daunting task that requires a lot of effort and resources. In contrast, images with only image-level supervision(\emph{i.e.} category labels) could be easily collected from existed dataset (\emph{e.g.} ImageNet~\cite{imagenet2009deng}) or Internet search engine. To this end, we put our concentration on investigating how to utilize these image-level annotated data to improve the performance of long-tailed object detection.



In this paper, we propose a novel long-tailed object detection framework named \textbf{CLIS} (multi-task \textbf{C}ollaborative \textbf{L}earning with \textbf{I}mage-level \textbf{S}upervision), which incorporates additional image-level supervision into the learning of object detectors in a multi-task collaborative way. As demonstrated in Fig.~\ref{fig:overlook}, there are two main tasks in our framework: an object detection task (consisting of an \textit{instance-classification} task and a \textit{localization} task) and an \textit{image-classification} task. They are responsible for the two types of supervision, respectively. Since the major cause of the performance degradation for long-tailed object detectors is the inaccurate prediction of the instance-classification task~\cite{simcal2020wang}, CLIS mainly focuses on improving the performance for this task with the help of extra knowledge collaboratively learned by the image-classification task.

To achieve this goal, three key components are designed in our framework.
Firstly, we propose to adopt the task-specialized sub-networks to learn specific representations of different tasks. It disentangles the features of the localization task and the two classification tasks, making them have a clear division of labor.
Then, a siamese sub-network is introduced for the image-classification task, which brings its knowledge to the instance-classification task by parameter sharing. This siamese structure enriches feature representations of the instance-classification task, which indeed enhances the long-tailed object detection ability. 
Finally, due to the two classification tasks receiving data from two different types of supervision, there is a feature gap between them during the knowledge sharing, preventing image-level supervision from making its best in our framework. To address this problem, we propose a contrastive learning regularization method to bridge the feature gap between the two classification tasks, keeping their consistency through a contrastive loss.
By the synergy of these components, CLIS could collaboratively learn knowledge across multi-tasks, taking full advantage of additional data to improve detection performance.

Extensive experiments are conducted to demonstrate the effectiveness of our proposed method. On the challenging LVISv1.0~\cite{lvis2019gupta} benchmark with the image-level supervision from the ImageNet-22k~\cite{imagenet2009deng} dataset, our approach achieves an overall AP of 31.1, bringing significant improvement for rare categories and establishing a new state-of-the-art. Experimental results for other tasks, \emph{e.g.} instance segmentation, also demonstrate the generalization ability of our method. Meanwhile, although training with additional data, our proposed framework introduces negligible computational cost during inference, making it a practice method in realistic long-tailed scenarios.


\section{Related Work}
\label{sec:related_work}
\subsection{Long-tailed Object Detection}
Long-tailed object detection is a challenging vision task receiving growing attention today. General solutions for this task are data re-sampling~\cite{lvis2019gupta,chang2021_sample_object_level,simcal2020wang} and cost-sensitive learning~\cite{eql2020tan,eqlv22021tan,wang2021adaptive,efl2022li,seesawloss2021wang,pan2021model,he2022relieving} that re-balance the contribution of different categories or instances to achieve a balanced training status. Decoupled training methods~\cite{crt2019kang,simcal2020wang,bags2020li,disalign2021zhang} decouple the learning of representation and classifier into two separated stages to address the classifier imbalance problem. Besides, there are also many other methods that make their effort on incremental learning~\cite{lst2020hu}, causal inference~\cite{causal_tde2020tang}, and so on. Nevertheless, all these methods try to solve the long-tailed problem given the training data with only instance-level supervision. However, the scarce instance number of rare categories prevents the model from learning discriminative features for classification. In contrast, our method makes use of extra image-level annotations to improve the classification ability of the long-tailed object detectors.

\subsection{Object Detection with Image-Level Supervision}
There are plenty of works that adopt image-level supervision in the object detection task. Weakly-supervised object detection (WSOD)~\cite{bilen2016wsddn,diba2017weakly,tang2017ocir,tang2018pcl,cinbis2016weakly,ren2020instance,zeng2019wsod2} trains object detectors from images with only image-level supervision, formulating the task as multiple instance learning (MIL) problem. Due to the lack of location information, the accuracy of these methods is far behind that of supervised object detectors, especially in some complex scenes. 
Semi-supervised object detection~\cite{jeong2019consistency,sohn2020stac,zhou2021instant,liu2021unbiased} trains the instance-level supervision data together with unlabeled images. And the Semi-supervised WSOD methods~\cite{redmon2017yolo9000,gao2019note_rcnn,uijlings2018revisiting,zhong2020boosting,DLWL2020Ramanathan,zhang2021mosaicos,zhou2022detecting} learn detectors with additional image-level supervision, which have the similar setting to our method. Among them, DLWL~\cite{DLWL2020Ramanathan} and MosaicOS~\cite{zhang2021mosaicos} improve the performance of low-shot categories with image-level supervision either by a linear program constraint or a multi-stage self-training framework. However, all these methods learn the image-level annotations as the weakly supervision to generate the boxes for the detection task, which are heavily dependent on the accuracy of the pseudo-label generation algorithm and may introduce too much noise. Besides, the recently proposed method Detic~\cite{zhou2022detecting} trains the classifier of the detector from the data coming from the two types of supervision which could be viewed as a multi-task training process. However, it does not take into account the feature entanglement of different tasks, let alone bridge the feature gaps among them. In this paper, we treat the learning of the two types of supervision as multi-tasks with a clear division. Based on our proposed framework, different tasks could be trained collaboratively, taking full advantage of image-level supervision to improve the long-tailed object detection performance.

\subsection{Multi-Task Learning and Collaborative Learning}
Multi-task learning approaches~\cite{zamir2020robust,zamir2018taskonomy,touvron2021training,zhang2021survey,bhattacharjee2022mult} learn to predict multiple outputs for a series of tasks jointly by a shared feature encoder/representation. They aim to improve the performance of all tasks by knowledge sharing between different tasks. However, in this work, our framework mainly focuses on learning the tasks related to long-tailed object detection, while the task for image-level supervision is utilized to bring its knowledge to improve the performance of detectors. Collaborative learning methods~\cite{qiao2018deep,niu2019multi,zhang2018deep,guo2021long} are usually applied between networks to learn them collaboratively for better feature representation or feature consistency. In this work, we formulate the collaboration by a siamese sub-network and a contrastive learning regularization method for better utilizing the image-level supervision.

\section{Methodology}
\subsection{Preliminary}
\label{subsec:preliminary}

In long-tailed object detection, there is a dataset $\mathcal{D}_d = \{(\boldsymbol{x}_d, \boldsymbol{y}_d)\}$\footnote{This full representation of the set is $\{(\boldsymbol{x}^1_d, \boldsymbol{y}^1_d), \dots, (\boldsymbol{x}^n_d, \boldsymbol{y}^n_d)\}$ with $n$ elements. For neatness, we simplify it and omit its superscripts. All following sets of this paper are the same.} with instance-level annotations. For each image $\boldsymbol{x}_d$, it contains a set of instances $\boldsymbol{y}_d = \{(\boldsymbol{b}, c)\}$, where $\boldsymbol{b} = (b_{cx}, b_{cy}, b_{w}, b_{h})$ is the location and $c \in C_d$ is the category label of an instance. $C_d$ is the set containing all categories in $\mathcal{D}_d$ that exhibit the long-tailed distribution. 

In our method, we introduce an extra dataset $\mathcal{D}_i = \{(\boldsymbol{x}_i, \boldsymbol{y}_i)\}$ with image-level supervision to alleviate the instance-scarce problem of tail categories. In $\mathcal{D}_i$, each image $\boldsymbol{x}_i$ is labeled by $\boldsymbol{y}_i = c$, where $c \in C_i, (C_i \subset C_d)$. The image-level label $\boldsymbol{y}_i$ indicates that there is at least one instance of category $c$ in the image $\boldsymbol{x}_i$. Our goal is to leverage these image-level annotations to improve the performance of long-tailed object detectors.

\begin{figure*}[ht]
  \begin{center}
  \includegraphics[width=0.98\linewidth]{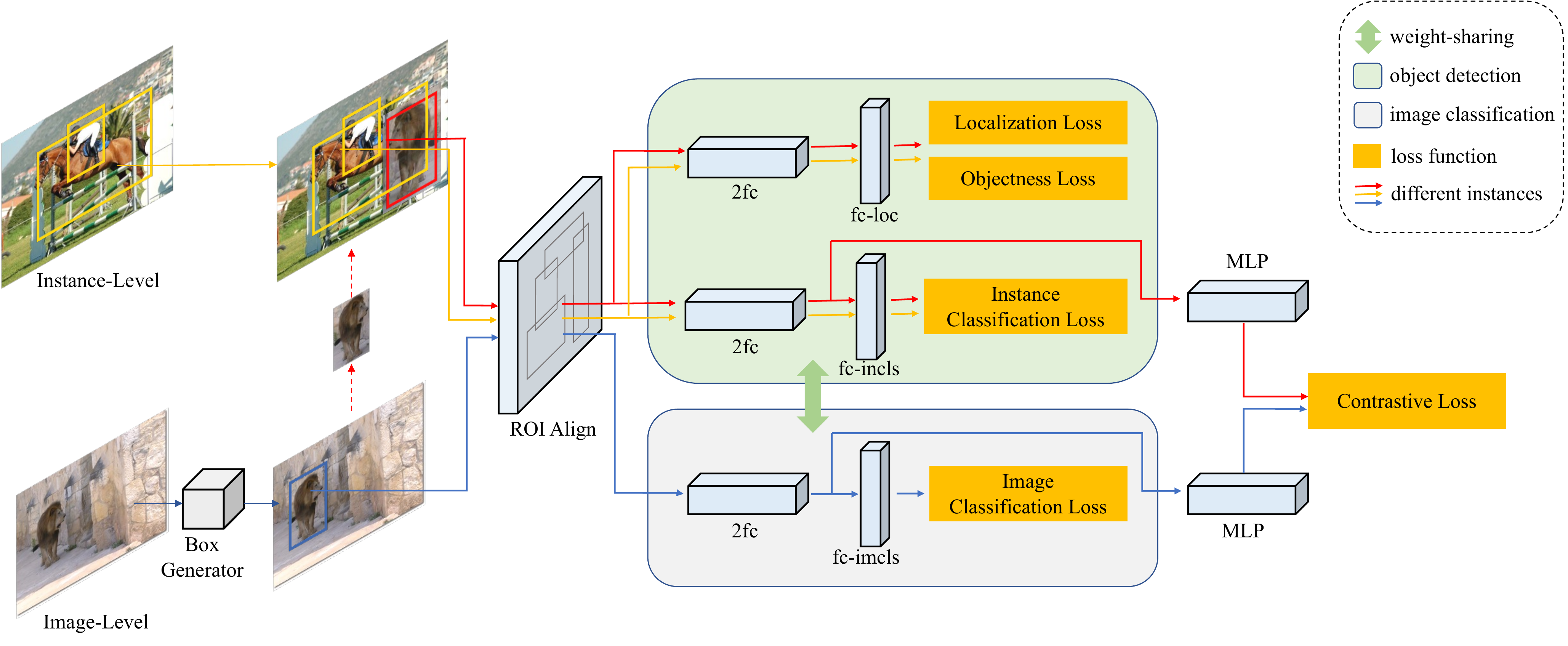}
  \end{center}
  \caption{Training pipeline of our proposed framework. The localization task, the instance-classification task, and the image-classification task are learned by their task-specialized sub-networks. The siamese sub-network (\textit{2fc + fc-imcls}) shares its weight with the instance-classification sub-network (\textit{2fc + fc-incls}). And a contrastive learning method is applied to the two tasks. The synergy of these components enables our framework to take full advantage of the image-level supervision for better learning the long-tailed object detectors.}
  \label{fig:pipeline}
\end{figure*}

\subsection{Multi-Task Collaborative Learning}
\label{subsec:multi_task_leanring}

We propose the multi-task collaborative learning framework to train object detectors with the help of image-level supervision. The framework is built upon a standard detection pipeline Faster R-CNN~\cite{faster_rcnn2015ren}. As presented in Fig.~\ref{fig:pipeline}, three tasks are involved in our framework which are a localization task, an instance-classification task, and an image-classification task. During training, in each iteration, we compose a batch of images $(\mathcal{X}_d; \mathcal{X}_i)$ from both $\mathcal{D}_d$ and $\mathcal{D}_i$. For images $\mathcal{X}_d$ with instance-level supervision, we follow the standard detection recipe and train them by the instance-classification task and the localization task. For images $\mathcal{X}_i$ with image-level supervision, we learn them as an image-classification task. In the long-tailed scenario, the instance-level supervision could help the model learn to locate an object well, while the classification ability of tail categories is still limited~\cite{simcal2020wang}. Therefore, we mainly focus on learning the instance-classification task with the help of the extra knowledge brought by the image-classification task.

\subsubsection{Task-Specialized Sub-Network}


For each task in our framework, we construct a sub-network for it. All sub-networks have the same structure of \textit{2fc} and a linear layer to output the final predictions. Especially, we separate the shared part (\emph{i.e.,} \textit{2fc} in R-CNN head) of the instance-classification task and the localization task into two parts without weight sharing, disentangling feature representations of the two tasks. In this way, the features of different tasks could keep their task information individually with a more clear division of labor. Such a design also facilitates the subsequent knowledge sharing between the two different classification tasks.


\subsubsection{Siamese Sub-Network}

To learn different tasks collaboratively, we extract the features of $\mathcal{X}_i$ by the feature extractor (\emph{i.e.,} the backbone with FPN~\cite{fpn2017lin}) shared with the detection framework. Different from standard image classification that applies AVGPooling on the whole image features, we perform ROIAlign on a pre-defined region that is most likely belonging to the image annotated instance (see Section~\ref{subsec:experiment_settings})\footnote{Note that although the training process of the image-classification task contains the pre-defined boxes, it is different from the object detection task. For example, we do not generate any positive/negative samples (or anchors) or leverage these boxes to learn the localization task.}. 
\begin{figure}[ht]
  \begin{center}
  \includegraphics[width=0.8\linewidth]{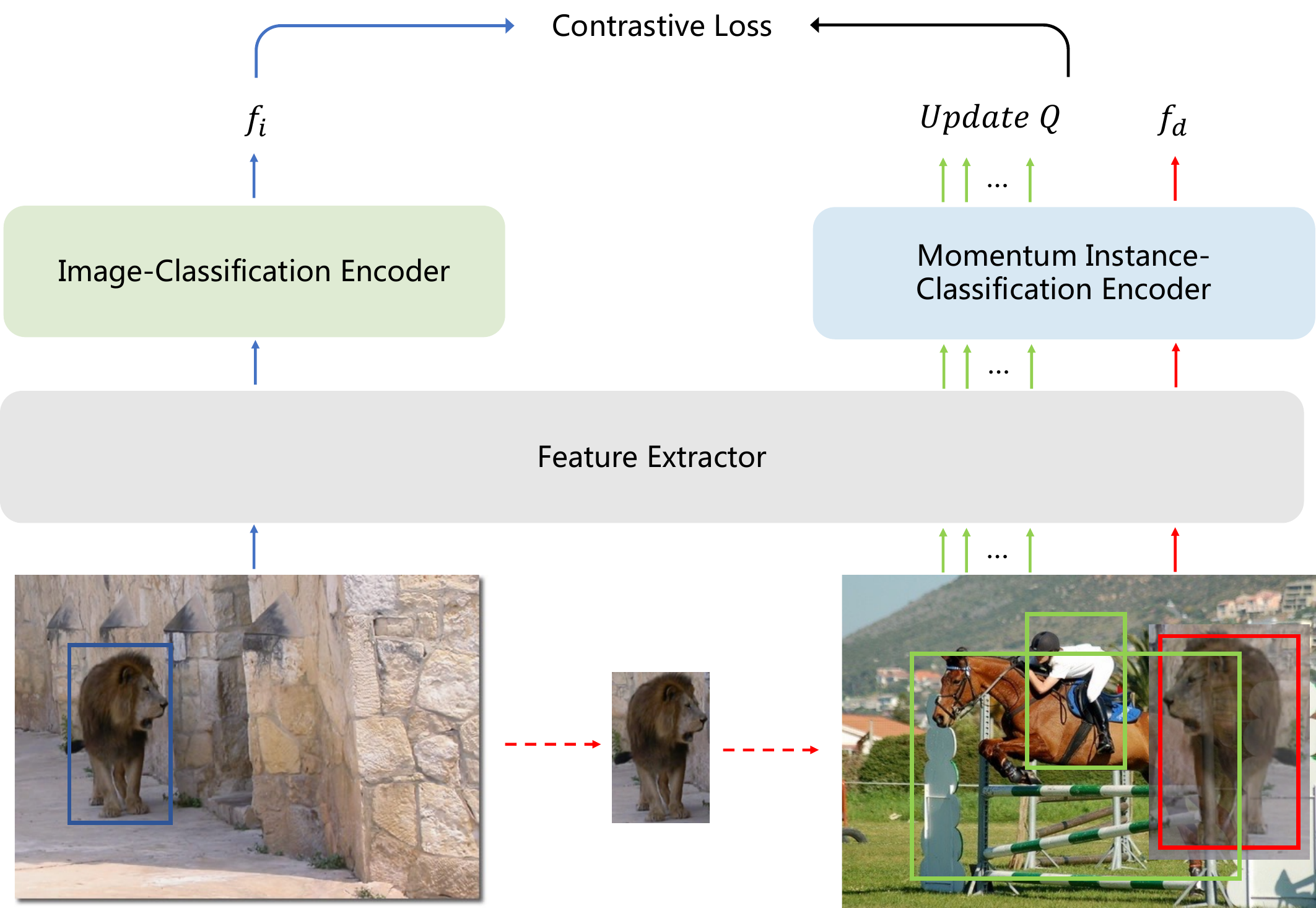}
  \end{center}
  \caption{An overview of the contrastive learning regularization. The blue and red colors indicate the two scenes of the same instance. And the green color is for other instances.}
  \label{fig:contrastive_method}
\end{figure}
This is because most images in $\mathcal{X}_i$ may contain multiple object instances of different categories in a complex background~\cite{zhang2021mosaicos}. 
Finally, a siamese sub-network for the image-classification task is introduced to learn the results of the predicted categories. It keeps the same structure (\emph{i.e.,} \textit{2fc} and \textit{fc-cls}) and parameters with the instance-classification sub-network. By weight sharing, the knowledge of image-level supervision in the siamese sub-network could enrich the feature representation of the instance-classification task for long-tailed object detectors.

\subsubsection{Contrastive Learning Regularization}

Even with the siamese sub-network, the feature representation of the two types of supervision may be totally different even for the same category. 
This is because the input data of the two classification tasks come from different scenes (\emph{i.e.,} an object detection scene and an image classification scene). 
These feature gaps are not conducive to knowledge sharing, thus preventing our framework from achieving excellent performance.
Especially, in the long-tailed situation, the gaps could be expanded further on tail categories because the model exhibits high uncertainty in them~\cite{zhang2021tade}. 

To solve this problem, we'd like to keep the feature consistency of two different scenes for the same instance.
As presented in Fig.~\ref{fig:contrastive_method}, for an instance in $\mathcal{X}_i$ (image classification scene), we could construct its counterpart in the object detection scene by picking it out and mixing it up into the image in $\mathcal{X}_d$. 
Then for each picked instance, we could obtain its feature representations from the two different scenes by training them with the image-classification task and the instance-classification task, respectively. 
A contrastive loss is adopted to maintain their feature consistency.
Following~\cite{he2020moco,chen2020mocov2}, we add an MLP projection layer onto the sub-network of each task to get the representation $f_i$ from the image-classification task and the representation $f_d$ from the instance-classification task.
A dictionary $Q$ is maintained as a queue of representation of other instances coming from both two tasks. We train $f_i$ with $f_d$ as its positive view and other representations in $Q$ as its negative views.
In this way, feature representations of the two tasks could be consistent for the same instance and keep diversity with other instances. The formula for the collaborative loss is:

\begin{equation}
  \mathcal{L}_{con} = -\mathrm{log} \frac{\mathrm{exp}(f_i \cdot f_d / \tau)}{\mathrm{exp}(f_i \cdot f_d / \tau) + \sum_{q \in Q} \mathrm{exp}(f_i \cdot q / \tau)}
\end{equation}

\noindent where $\tau$ is a temperature hyper-parameter. This contrastive loss serves as a feature regularization that ensures the learning of two tasks consistently, resulting in efficient usage of image-level supervision in our framework.

\subsubsection{Loss Formula}

Finally, we have the detection loss $\mathcal{L}_d$ for instance-level supervision learning, the classification loss $\mathcal{L}_i$ for image-level supervision learning, and the contrastive loss $\mathcal{L}_{con}$ for feature regularization. All losses are optimized jointly:

\begin{equation}
  \begin{aligned}
    \mathcal{L} &= \mathcal{L}_d + \alpha \mathcal{L}_{i} + \beta \mathcal{L}_{con} \\
    &= \mathcal{L}_{rpn} + \mathcal{L}_{incls} + \mathcal{L}_{loc} + \mathcal{L}_{obj} \\
    &\:\:\:\:\:\:+ \alpha \mathcal{L}_{imcls} + \beta \mathcal{L}_{con}
  \end{aligned}
  \label{eq:losses}
\end{equation}

\noindent where $\mathcal{L}_d$ consists of several terms which are the $\mathcal{L}_{rpn}$ from RPN, $\mathcal{L}_{incls}$ from the instance-classification task, and $\mathcal{L}_{loc}$ from the localization task. Similar to~\cite{bags2020li,eqlv22021tan}, we add an objectiveness loss $\mathcal{L}_{obj}$ on the localization sub-network to reduce the number of false positive predictions. 
For $\mathcal{L}_{i}$, it directly calculates the loss $\mathcal{L}_{imcls}$ from the image-classification task by the siamese sub-network. $\alpha$ and $\beta$ are hyper-parameters used to balance the loss contribution of each module. We will show later that it is important to set them properly to achieve good results.

\subsubsection{Model Inference}

During inference, we directly predict the result through the instance-classification task and the localization task. The objectiveness estimation score is multiplied by the classification score to get the final estimated probability. Since the image-level annotated data is not involved in the inference process, our framework introduces negligible computational costs to the inference time, demonstrating the efficiency of our method (more details could be found in Section~\ref{subsubsec: data_efficiency}).

\section{Experiments}
\label{sec:experiments}

\subsection{Experimental Settings}
\label{subsec:experiment_settings}
\subsubsection{Dataset Setup} We adopt the challenging LVISv1.0~\cite{lvis2019gupta} as the dataset $\mathcal{D}_d$ with instance-level supervision. LVIS is a large vocabulary dataset for both object detection and instance segmentation. It contains 1203 categories, following long-tailed distribution. Each category has a unique id of WordNet~\cite{miller1995wordnet}. There are total 100k images with 1.3M instances in the \texttt{train} set and 20k images in the \texttt{val} set. Except for the widely used AP metric (average precision of boxes prediction across IoU thresholds from 0.5 to 0.95), we also report AP\textsubscript{r} (rare categories with 1-10 images), AP\textsubscript{c} (common categories with 11-100 images),  and AP\textsubscript{f} (frequent categories with $>$100 images). A subset of ImageNet-22k~\cite{imagenet2009deng} is used as the dataset $\mathcal{D}_i$ with image-level supervision. ImageNet-22k contains 21842 categories which are also associated with WordNet ids. By matching these ids with LVIS, we finally collect 1,237,737 images from the \texttt{train} split as our dataset $\mathcal{D}_i$, with a total of 997 categories overlapped. The image numbers of $\mathcal{D}_i$ are about 10x bigger than $\mathcal{D}_d$ with a relatively balanced category distribution which could provide sufficient information to help the learning of tail categories.

\begin{table*}
    \centering
    \setlength\tabcolsep{16pt}
    \caption{Comparison with the state-of-the-art methods. \textsuperscript{\textdaggerdbl} indicates that the results are copied from~\cite{zhang2021mosaicos}. `IN' and `G' means ImageNet-22k and Google Images. The rebalance-based methods are trained by a 2x scheduler with RFS~\cite{lvis2019gupta} while~\cite{zhang2021mosaicos} and CLIS are trained by 1x. Note that BAGS~\cite{bags2020li} has an additional 1x schedule fine-tuning stage with the class-balanced sampler.}
    \begin{tabular}{l | l | c c | c c c c}
      \toprule
      backbone & method & data & scheduler & AP & AP\textsubscript{r} & AP\textsubscript{c} & AP\textsubscript{f} \\
      \midrule
      \multirow{7}{*}{ResNet-50} & Faster R-CNN~\cite{faster_rcnn2015ren} & - & 2x & 24.1 & 14.7 & 22.2 & 30.5  \\
      ~ & EQLv2~\cite{eqlv22021tan} & - & 2x & 25.5 & 16.4 & 23.9 & 31.2 \\
      ~ & BAGS~\cite{bags2020li} & - & 2x & 26.0 & 17.2 & 24.9 & 31.1 \\
      ~ & Seesaw Loss~\cite{seesawloss2021wang} & -  & 2x & 26.4 & 17.5 & 25.3 & 31.5 \\
      ~ & EFL~\cite{efl2022li} & -  & 2x & 27.5 & 20.2 & 26.1 & \textbf{32.4} \\
      ~ & MosaicOS\textsuperscript{\textdaggerdbl}~\cite{zhang2021mosaicos} & IN+G & 1x & 23.9 & 15.5 & 22.4 & 29.3 \\
      ~ & \textbf{CLIS (ours)} & IN & 1x & \textbf{29.2} & \textbf{24.4} & \textbf{28.6} & 31.9 \\
      \midrule
      \multirow{6}{*}{ResNet-101} & Faster R-CNN~\cite{faster_rcnn2015ren} & - & 2x & 25.7 & 15.1 & 24.1 & 32.0 \\
      ~ & EQLv2~\cite{eqlv22021tan} & - & 2x & 26.9 & 18.2 & 25.4 & 32.4 \\
      ~ & BAGS~\cite{bags2020li} & - & 2x & 27.6 & 18.7 & 26.5 & 32.6 \\
      ~ & Seesaw Loss~\cite{seesawloss2021wang} & - & 2x & 27.8 & 18.7 & 27.0 & 32.8 \\
      ~ & EFL~\cite{efl2022li} & - & 2x & 29.2 & 23.5 & 27.4 & \textbf{33.8} \\
      ~ & \textbf{CLIS (ours)} & IN & 1x & \textbf{31.1} & \textbf{25.2} & \textbf{30.9} & \textbf{33.8} \\
      \bottomrule
    \end{tabular}
    \label{tab:main_result}
\end{table*}


\subsubsection{Implementation Details.} We implement our method based on the MMDetection~\cite{chen2019mmdetection} framework. All networks are trained 90k iterations by the repeat factor sampler~\cite{lvis2019gupta} using the SGD algorithm with a momentum of 0.9 and a weight decay of 0.0001. The initial learning rate is set as 0.02. During the training phase, for images $\mathcal{X}_d$ from the LVIS dataset $\mathcal{D}_d$, scale jitter and horizontal flipping are adopted as the data augmentation. The batch size of $\mathcal{X}_d$ is set as $\mathcal{B}_d$, where $\mathcal{B}_d = 16$ on 16 GPUs. For images $\mathcal{X}_i$ from the ImageNet-22k dataset $\mathcal{D}_i$, we augment them by random scaling, horizontal flipping, cutout~\cite{cutout2017devries}, and mosaic~\cite{bochkovskiy2020yolov4}. Each mosaic image is synthesized by four randomly picked images and resized to $448 \times 448$. We set the 
batch size of $\mathcal{X}_i$ as $\mathcal{B}_i$, where $\mathcal{B}_i = s\mathcal{B}_d$. Typically, we have $s = 16$ which means that one detection image is trained along with $s$ classification images. Among the $s$ images, we pick $t$ instances for contrastive learning regularization. We have $t=2$ by default. We train $\mathcal{X}_d$ and $\mathcal{X}_i$ jointly with the ImageNet-1K pre-trained Resnet~\cite{resnet2016he} as the backbone. For the contrastive learning regularization method in our framework, the MLP is set as a \textit{fc} with the output dim 128. The size of the dictionary $Q$ and the temperature $\tau$ are set as 115712 and 0.2, respectively, followed by~\cite{li2021self}. And a momentum update mechanism is applied to the instance-classification sub-network and its MLP to generate features in $Q$. For our proposed CLIS, hyper-parameters $\alpha$ and $\gamma$ are set as 0.1 and 0.05, respectively. During inference, we evaluate the images from the \texttt{val} split of $\mathcal{D}_d$ with a standard detection evaluation setting. No testing time augmentation is used. Following~\cite{lvis2019gupta}, we select the top 300 boxes with confidence scores greater than 0.0001 per image as the final detection results. 


\subsubsection{Pre-defined Region Generation} 
As we described above, for each image of $\mathcal{X}_i$, we need to generate a pre-defined instance region to indicate where the part most likely to belong to the annotation category is. We generate these regions by a pre-trained detector on $\mathcal{D}_d$. Basically, we only need one region of each image but the pre-trained detector would output a list of detection bounding boxes from different categories. Directly picking the box with the highest prediction score may not be a good choice because tail categories in $\mathcal{D}_d$ usually have low scores due to the long-tailed distribution.
In this paper, we observe that the detector could rank the proposals accurately if we provide the image-level label of the images (a similar phenomenon is described in~\cite{dave2021devilap}). Then we follow the category-rank-first rule to generate the pre-defined region of the instance. For example, if an image of $\mathcal{X}_i$ is annotated with a category \textit{eagle}, then we will pick the box with the highest score of \textit{eagle} among the prediction results as the pre-defined region. In our methods, all  pre-defined regions are generated offline with a baseline model (1x, RFS) as the pre-trained detector. 

\subsection{Benchmark Results}

To demonstrate the strength of our proposed method, we compare our approach with other works that report state-of-the-art performance. Note that, to obtain a converged training status, we train the rebalance-based methods by a 2x scheduler. Whereas for the methods using additional image-level supervision, 1x is sufficient. As demonstrated in Table~\ref{tab:main_result}, with the ResNet-50-FPN backbone, our CLIS framework significantly outperforms the baseline Faster R-CNN~\cite{faster_rcnn2015ren} method by 5.1 AP, achieving an overall AP of 29.2. More importantly, CLIS significantly improves the performance of rare categories by 9.7 AP, indicating the effectiveness of our proposed method in solving the long-tailed object detection problem. Compared with the mainstream rebalance-based approaches, CLIS equipped with a simple CE loss outperforms EQLv2~\cite{eqlv22021tan}, BAGS~\cite{bags2020li}, Seesaw Loss~\cite{seesawloss2021wang}, and EFL~\cite{efl2022li} by 3.7 AP, 3.2 AP, 2.8 AP, and 1.7 AP, respectively. And when compared with MosaicOS~\cite{zhang2021mosaicos} which also utilizes image-level supervision to train the detector, our CLIS framework surpasses it by a large margin with 5.3 AP improvement. More experiments about the comparison with the approaches which also make use of the image-level annotations could be found in Section~\ref{subsec:model_analysis}. 

\begin{figure}
  \begin{center}
  \includegraphics[width=0.98\linewidth]{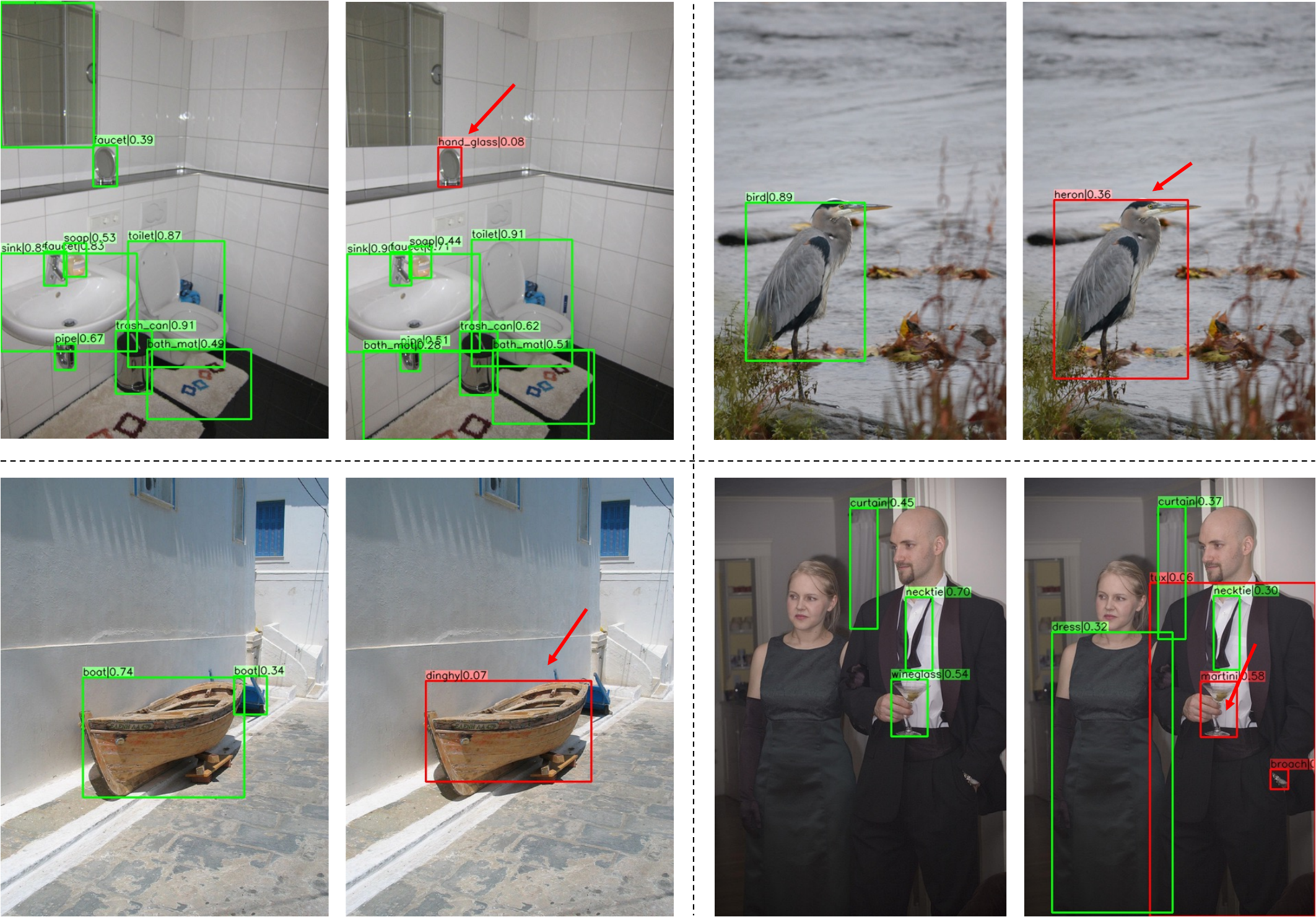}
  \end{center}
  \caption{Results on LVISv1.0 \texttt{val} split. For each pair of images, the left one is from the baseline model and the right one is from the CLIS framework. Green and red boxes indicate the frequent and rare categories, respectively. The red arrows show the clear improvement of our approach.}
  \label{fig:results}
\end{figure}

On the larger ResNet-101-FPN backbone, our proposed method consistently achieves high performance on both AP and AP\textsubscript{r}, surpassing Faster R-CNN~\cite{faster_rcnn2015ren} by 5.4 AP and 10.1 AP. Without bells and whistles, CLIS achieves an overall AP of 31.1, outperforming all existing approaches. The state-of-the-art performance demonstrates the power and advantages of our framework in leveraging image-level supervision to help the learning of long-tailed object detection. Additionally, we show some qualitative analysis in Fig.~\ref{fig:results}, CLIS outperforms the baseline with more accurate classification results of rare categories.

\begin{table}
  \centering
  \setlength{\tabcolsep}{8pt}
  \caption{Ablation study of each component of CLIS: TSS (task-specialized sub-networks), SS (siamese sub-network), CLR (contrastive learning regularization), and ILS (image-level supervision).}
  \begin{tabular}{l|cccc}
    \toprule
    method & AP & $AP_r$ & $AP_c$ & $AP_f$ \\
    \midrule
    \textbf{CLIS} & \textbf{29.2} & \textbf{24.4} & \textbf{28.6} & 31.9 \\
    w/o (TSS) & 25.9 & 20.3 & 25.0 & 29.5 \\
    w/o (SS) & 28.3 & 21.7 & 27.6 & \textbf{32.0} \\
    w/o (CLR) & 28.2 & 20.5 & 27.8 & \textbf{32.0} \\
    w/o (ILS) & 25.2 & 14.6 & 24.0 & 31.2 \\
    w/o (TSS + ILS) & 22.0 & 8.9 & 20.8 & 29.1 \\
    \bottomrule
  \end{tabular}
  \label{tab:ablation_study_components}
\end{table}

\subsection{Ablation Studies}

\subsubsection{Ablation of the component}

To verify the effectiveness of each design choice in the proposed method, we conduct a comprehensive ablation study here. We mainly study the three proposed key components in our framework, which are the Task-Specialized Sub-networks (TSS), the Siamese Sub-network(SS), and the Contrastive Learning Regularization method(CLR). Besides, we also investigate the overall influence of Image-Level Supervision (ILS). We show the effect of each component by taking it away from the framework. Here we present the detailed experimental configurations for different components:

\textit{w/o (TSS):} Learning the framework without disentangling features of different tasks. In other words, all tasks share the same sub-network.

\textit{w/o (SS):} Learning the instance-classification task and the image-classification task without weight sharing. The result is all tasks keep their specialized sub-networks.

\textit{w/o (CLR):} Learning the instance-classification task and the image-classification task without the contrastive learning regularization method.

\textit{w/o (ILS):} Learning the framework without image-level supervision.

\textit{w/o (TSS + ILS):} Learning the framework without task-specialized sub-networks and image-level supervision. In other words, the model is the baseline Faster R-CNN framework.

As demonstrated in Table~\ref{tab:ablation_study_components}, if trained without the task-specialized sub-networks, the performance of CLIS will degrade a lot by 3.3 AP. It is because, without this component, different tasks in our framework are learned in feature entanglement, negatively impacting the overall performance. And when trained without the siamese sub-network, the instance-classification task and the image-classification task will have no knowledge sharing between their sub-networks, resulting in a performance decrease from 29.2 AP to 28.3 AP. It is worth noting that even without the siamese sub-network, the image-level supervision could still bring a 3.1 AP improvement (from 25.2 to 28.3). Such an improvement mainly comes from the shared backbone, disentanglement features, and contrastive learning regularization of different tasks. This result also indicates the value of image-level supervision in our multi-task framework. Meanwhile, if the model is trained without the contrastive learning regularization method, the feature gaps from different types of supervision may disturb the knowledge-sharing process, decreasing the performance by 1.0 AP. With the synergy of these components, CLIS dramatically improves the performance of the baseline method from 22.0 AP to 29.2 AP, bringing the best of image-level supervision to long-tailed object detection.


\begin{table}
    \centering
    \setlength\tabcolsep{10pt}
    \caption{Ablation study of the hyper-parameter $\alpha$.}
    \begin{tabular}{c | c c c c}
       \toprule
       $\alpha$ & AP & AP\textsubscript{r} & AP\textsubscript{c} & AP\textsubscript{f} \\
       \midrule
       0 & 25.2 & 14.6 & 24.0 & 31.2 \\
       0.01 & 26.1 & 16.6 & 25.2 & 31.3 \\
       0.05 & 27.3 & 17.5 & 27.1 & 31.8 \\
       \textbf{0.1} & \textbf{28.2} & \textbf{20.5} & \textbf{27.8} & \textbf{32.0} \\
       0.2 & 28.0 & 19.5 & 27.9 & 31.9\\
       0.5 & 26.5 & 17.8 & 26.5 & 30.4 \\
      \bottomrule
    \end{tabular}
    \label{tab:ablation_alpha}
\end{table}


\subsubsection{Ablation of the hyper-parameters} 
\label{subsubsec: hypers}
$\alpha$ and $\beta$ are two hyper-parameters in CLIS that balance the contribution of the image-classification task and the contrastive learning regularization method. Because the contrastive learning regularization method must rely on the sub-networks of both instance-classification task and image-classification task, we can not analyze $\beta$ individually without $\alpha$. Therefore, we first study the impact of $\alpha$ in Table~\ref{tab:ablation_alpha}. The baseline model is set as the framework trained without image-level supervision ($\alpha=0$ and $\beta=0$). It could be observed from the table that $\alpha=0.1$ is a good balance factor for the image-classification task, achieving 28.2 AP for long-tailed object detection. A too small value of $\alpha$ like 0.01 can not take full advantage of the image-level data while a too large value of $\alpha$ like 0.5 may impair the original feature representations of the instance-classification task, resulting in marginal improvements. Then we fix the value of $\alpha=0.1$ and investigate the effect of $\beta$. As shown in Table~\ref{tab:ablation_beta}, setting the $\beta$ a positive value will further improve the performance of the detector, demonstrating the strength of the contrastive learning regularization method. And $\beta=0.05$ is a suitable factor that achieves an overall AP of 29.2. A large value of $\beta$ like 0.1 may lead to too strict a constraint, resulting in a NaN (Not a Number) problem. Finally, we set the $\alpha=0.1$ and $\beta=0.05$ to balance the training status, by default.

\begin{table}
    \centering
    \setlength\tabcolsep{10pt}
    \caption{Ablation study of the hyper-parameter $\beta$.}
    \begin{tabular}{c | c c c c}
       \toprule
       $\beta$ & AP & AP\textsubscript{r} & AP\textsubscript{c} & AP\textsubscript{f} \\
       \midrule
       0 & 28.2 & 20.5 & 27.8 & \textbf{32.0} \\
       0.01 & 28.9 & 22.5 & \textbf{28.8} & 31.9 \\
       \textbf{0.05} & \textbf{29.2} & \textbf{24.4} & 28.6 & 31.9 \\
       0.08 & 28.7 & 21.8 & 28.6 & \textbf{32.0} \\
       0.1 & NaN & - & - & - \\
      \bottomrule
    \end{tabular}
    \label{tab:ablation_beta}
\end{table}

\begin{table}
    \centering
    \setlength\tabcolsep{8pt}
    \caption{Ablation study of the data efficiency $s$.}
    \begin{tabular}{c | c c c c | c}
        \toprule
        $s$ & AP & AP\textsubscript{r} & AP\textsubscript{c} & AP\textsubscript{f} & train time\\
        \midrule
        0 & 25.2 & 14.6 & 24.0 & 31.2 & 0.249s \\
        4 & 27.7 & 19.5 & 27.5 & 31.5 & 0.275s \\
        8 & 27.8 & 18.5 & 27.7 & 31.9 & 0.296s \\
        \textbf{16} & \textbf{28.2} & \textbf{20.5} & 27.8 & \textbf{32.0} & 0.381s \\
        32 & \textbf{28.2} & 20.0 & \textbf{27.9} & \textbf{32.0} & 0.416s \\
        \bottomrule
    \end{tabular}
    \label{tab:ablation_batch_s}
\end{table}

\subsubsection{Ablation of the data efficiency} 
\label{subsubsec: data_efficiency}
CLIS shares the knowledge of the additional image-level supervision with the learning of instance-level supervision by the siamese sub-network of the image-classification task and the contrastive learning regularization method. 
Here we study the performance with respect to different sizes $s$ of image-level supervision in the image-classification task and different numbers $t$ of the picked instances in the contrastive learning regularization method. We first study the influence of $s$ and set the baseline model as $s=0$ and $t=0$. As demonstrated in Table~\ref{tab:ablation_batch_s}, a small size of $s$ like 4 (4 image-level data trained with 1 instance-level data), could improve the performance from 25.2 AP to 27.7 AP. And a larger value of $s$ could yield better performance which has already been proved in many image-classification tasks~\cite{imagenet2009deng,resnet2016he}. We choose the $s=16$ as the default setting because it keeps a good trade-off between the training costs and accuracy. Note that when we train the detection data $\mathcal{D}_d$ for one epoch, setting the $s=16$ will also train the image-level data $\mathcal{D}_i$ for about one epoch because the image numbers of $\mathcal{D}_i$ are about 10x bigger than $\mathcal{D}_d$. Then we fix the size of $s$ and sample $t$ instance from them for the contrastive learning regularization method. Table~\ref{tab:ablation_batch_t} shows $t$ with 1x or 2x larger than the instance-level annotated data is a proper setting that could achieve a promising improvement by about 1.0 AP. While a larger value of $t$ like 4 brings little improvement, indicating that too strict constraints may limit the model diversity. By default, we set $s=16$ and $t=2$ for all other experiments in this paper. What's more, we show the training times of our framework with different data scales under the Nvidia V100 GPUs. It could be observed that even if we bring 16x image-level data into each batch ($s=16$ and $t=2$), the increase in training time is less than 2x (from 0.249s to 0.479s), indicating the efficiency of our method in leveraging the image-level data. It is worth noting that the testing time for all experiments is exactly the same as that of the baseline model.

\begin{table}
    \centering
    \setlength\tabcolsep{8pt}
    \caption{Ablation study of the data efficiency $t$.}
    \begin{tabular}{c | c c c c | c}
        \toprule
        t & AP & AP\textsubscript{r} & AP\textsubscript{c} & AP\textsubscript{f} & train time\\
        \midrule
        0 & 28.2 & 20.5 & 27.8 & 32.0 & 0.381s\\
        1 & 28.8 & 21.8 & \textbf{28.6} & \textbf{32.1} & 0.403s\\
        \textbf{2} & \textbf{29.2} & \textbf{24.4} & \textbf{28.6} & 31.9 & 0.479s\\
        4 & 28.3 & 22.3 & 28.1 & 31.3 & 0.531s\\
        \bottomrule
    \end{tabular}
    \label{tab:ablation_batch_t}
\end{table}

\begin{table*}
    \centering
    \setlength\tabcolsep{10pt}
    \caption{Comparison with other methods which also utilize image-level supervision in long-tailed object detection. \textsuperscript{\textdagger} indicates that the results are copied from \cite{DLWL2020Ramanathan}. And \textsuperscript{\textdaggerdbl} are copied from \cite{zhang2021mosaicos}. `IN' and `G' means ImageNet-22k and Google Images.}
    \begin{tabular}{l | c c| l l l l }
        \toprule
        method & LVIS version & Data  & AP & AP\textsubscript{r} & AP\textsubscript{c} & AP\textsubscript{f} \\
        \midrule
        Faster R-CNN\textsuperscript{\textdagger} & v0.5 & -  & 21.9 & 10.8 & - & - \\
        DLWL\textsuperscript{\textdagger}~\cite{DLWL2020Ramanathan} & v0.5 & YFCC-100M & 22.1 (+0.2) & 14.2 (+3.4) & - & - \\
        \midrule
        Faster R-CNN\textsuperscript{\textdaggerdbl} & v0.5 & - & 23.2 & 12.6 & 22.4 & 28.3 \\
        Self-Training\textsuperscript{\textdaggerdbl}~\cite{zoph2020rethinking_pretrain_selftrain} & v0.5 & IN & 22.7 (-0.5) & 14.5 (+1.9) & 21.4 (-1.0) & 27.6 (-0.7) \\
        MosaicOS\textsuperscript{\textdaggerdbl}~\cite{zhang2021mosaicos} & v0.5 & IN & 24.8 (+1.6) & 19.7 (+7.1) & 23.4 (+1.0) & 28.4 (+0.1) \\
        MosaicOS\textsuperscript{\textdaggerdbl}~\cite{zhang2021mosaicos} & v0.5 & IN+G & 25.0 (+1.8) & 20.3 (+7.7) & 23.9 (+1.5) & 28.3 (+0.0) \\
        \midrule
        Faster R-CNN & v0.5 & - & 24.2 & 13.3 & 23.3 & 29.7 \\
        \textbf{CLIS} & v0.5 & IN & \textbf{30.4 (+6.2)} & \textbf{25.5 (+12.2)} & \textbf{30.5 (+7.2)} & \textbf{32.1 (+2.4)} \\
        \midrule
        Faster R-CNN\textsuperscript{\textdaggerdbl} & v1.0 & - & 22.0 & 10.6 & 20.1 & 29.2 \\
        Unbiased Teacher~\cite{liu2021unbiased} & v1.0 & IN & 19.2 (-2.8) & 6.6 (-4.0) & 17.4 (-2.7) & 26.8 (-2.4) \\
        MosaicOS\textsuperscript{\textdaggerdbl}~\cite{zhang2021mosaicos} & v1.0 & IN+G & 23.9 (+1.9) & 15.5 (+4.9) & 22.4 (+2.3) & 29.3 (+0.1) \\
        \midrule
        Faster R-CNN & v1.0 & - & 22.0 & 8.9 & 20.8 & 29.1 \\
        \textbf{CLIS} & v1.0 & IN & \textbf{29.2 (+7.2)} & \textbf{24.4 (+15.5)} & \textbf{28.6 (+7.8)} & \textbf{31.9 (+2.8)} \\
        \bottomrule 
    \end{tabular}
    \label{tab:main_results}
\end{table*}

\subsection{Model Analysis}
\label{subsec:model_analysis}

\subsubsection{Comparison with Methods Using Image-Level Data}

To show the effectiveness of our proposed method, we compare our approach with other works that also utilize image-level supervision to improve long-tailed object detection. Besides the LVISv1.0, we also report the experimental results on the challenging LVISv0.5 benchmark\footnote{Note that LVISv0.5 contains 1230 categories and there are 1023 overlapping categories between the LVISv0.5 and ImageNet-22k based on WordNet ids. All models are trained by 90k iterations. Other settings are the same with LVISv1.0.}. As demonstrated in Table~\ref{tab:main_results}, on both datasets with ResNet-50 backbone, our proposed CLIS achieves state-of-the-art performance with 29.2 AP and 30.4 AP, surpassing the baseline Faster R-CNN~\cite{faster_rcnn2015ren} by 7.2 AP and 6.2 AP, respectively. For semi-supervised methods Self-Traing~\cite{zoph2020rethinking_pretrain_selftrain} and Unbiased Teacher~\cite{liu2021unbiased} which learn the additional images as unlabeled data, we can see that they do not improve much performance and even have negative effects on the performance. This is because these methods generate pseudo-labels heavily relying on the performance of the basic detectors, which will show inferior and noisy predictions in the long-tailed situation. Meanwhile, for semi-supervised WSOD methods which treat the additional images as weakly supervision to guide their pseudo-label generation process, we compare CLIS with two approaches: DLWL~\cite{DLWL2020Ramanathan} and MosaicOS~\cite{zhang2021mosaicos}. DLWL collects data from the large-scale YFCC-100M dataset with tags. On the LVISv0.5 dataset, it improves the AP of rare categories by 3.4 points but only brings marginal improvement for overall AP. MosaicOS uses the same ImageNet-22k subset as our method, with additional data from Google images. It improves the overall AP from 23.2 to 25.0 with 7.7 AP improvement for rare categories. Under the same settings, we train the baseline model with a stronger performance of 24.2 AP. And our CLIS method outperforms the baseline by 6.2 AP with 12.2 points of AP improvement for rare categories. On the LVISv1.0 dataset, CLIS consistently outperforms the baseline method and MosaicOS by 7.2 AP and 5.3 AP, respectively. Note that besides the great improvement for rare categories, our methods also perform well on the frequent categories with more than 2 AP improvement, which indicates that our multi-task learning framework could make good use of image-level supervision for all categories. 


\begin{table}
    \centering
    \setlength\tabcolsep{12pt}
    \caption{
      Influence of the size of image-level data.}
    \begin{tabular}{c| c c c c}
       \toprule
       size & AP & AP\textsubscript{r} & AP\textsubscript{c} & AP\textsubscript{f} \\
       \midrule
       0 & 25.2 & 14.6 & 24.0 & 31.2 \\
       10 & 27.6 & 21.3 & 26.8 & 31.3 \\
       50 & 28.7 & 23.1 & 28.1 & 31.9 \\
       \textbf{100} & \textbf{29.2} & \textbf{24.4} & \textbf{28.6} & \textbf{31.9} \\
      \bottomrule
    \end{tabular}
    \label{tab:effect_of_size}
  \end{table}

\subsubsection{Influence of Image-Level Data Size}

We study the effect of the size of the image-level supervision dataset $\mathcal{D}_i$. We randomly sample images from the 120k data in $\mathcal{D}_i$ by different ratios. The results are shown at Table~\ref{tab:effect_of_size}. With only 10\% data, whose size is roughly the same as that of LVIS data, the AP of rare categories can be improved significantly from 14.6 to 21.3. In LVIS, the rare categories only contain $< 10$ instances, making it hard for the detector to learn for rare categories. The 10\% additional data is more balanced and provides more training samples for them, thus improving the performance a lot. We find that the model achieves higher accuracy with more data. This demonstrates that the performance of long-tailed object detectors can be improved with large-scale image-level supervision data, making CLIS a practical method in a realistic scenario.

\subsubsection{Results on Instance Segmentation}

To further demonstrate the strength and generalization ability of our proposed method, we report the result of instance segmentation on the LVISv1.0 \texttt{val} split and compare it with state-of-the-art methods. All models are trained by a 2x scheduler and repeat factor sampler. When applying the CLIS to the Mask R-CNN R50~\cite{maskrcnn2017he} framework, we only utilize our proposed components on the bbox head without modification on the mask head. Note that the PCB~\cite{he2022relieving} result is based on the Seesaw Loss~\cite{seesawloss2021wang} and we report the Detic~\cite{zhou2022detecting} result without the CLIP~\cite{radford2021clip} classifier for a fair comparison. Compared with the rebalance-based methods, \emph{e.g.}, Seesaw Loss, and PCB, our proposed CLIS framework outperforms all of them by a simple cross-entropy loss. While compared to the methods that also leverage image-level supervision for training like MosaicOS~\cite{zhang2021mosaicos} and Detic, our method achieves the strongest AP of 29.3, surpassing them by a large margin. Especially for the Detic method which utilizes a similar multi-task training strategy, CLIS outperforms it by 4.2 AP, indicating the advantages of our proposed components.

\begin{table}
    \centering
    \setlength\tabcolsep{8pt}
    \caption{Instance segmentation comparison with the state-of-the-art methods on LVIS v1.0 \texttt{val}. All models are trained by a 2x scheduler with the repeat factor sampler. \textsuperscript{*} indicates that the results are copied from~\cite{he2022relieving}. And \textsuperscript{\textdaggerdbl} are copied from~\cite{zhang2021mosaicos} and~\cite{zhou2022detecting}}
    \begin{tabular}{l | c c c c c}
      \toprule
      method & AP\textsuperscript{\textit{m}} & AP\rlap{\textsuperscript{\textit{m}}}{\textsubscript{r}} & AP\rlap{\textsuperscript{\textit{m}}}{\textsubscript{c}} & AP\rlap{\textsuperscript{\textit{m}}}{\textsubscript{f}} & 
      AP\rlap{\textsuperscript{\textit{b}}}
      \\
      \midrule
      Mask R-CNN\textsuperscript{\textdaggerdbl}~\cite{maskrcnn2017he} & 23.7 & 13.5 & 22.8 & 29.3 & 24.7 \\
      Seesaw Loss\textsuperscript{*}~\cite{seesawloss2021wang} & 26.8 & 19.8 & 26.3 & 30.5 & 27.6 \\
      PCB\textsuperscript{*}~\cite{he2022relieving} & 27.2 & 19.0 & 27.1 & 30.9 & 28.1 \\
      MosaicOS\textsuperscript{\textdaggerdbl}~\cite{zhang2021mosaicos} & 24.5 & 18.3 & 23.0 & 28.9 & - \\
      Detic\textsuperscript{\textdaggerdbl}~\cite{zhou2022detecting} & 25.1 & 18.6 & - & - & - \\
      \textbf{CLIS (ours)} & \textbf{29.3} & \textbf{23.0} & \textbf{29.0} & \textbf{32.4} & \textbf{30.4} \\
      \bottomrule
    \end{tabular}
    \label{tab:instance_segmentation}
  \end{table}

\subsubsection{Pre-training or Joint-Training}

We compare our method with another way to make use of the additional image-level supervision data, which is the pre-training. Specifically, for pre-training, we first use the 120k images with image-level labels to pre-train the model and then fine-tune it with detection data. We follow the standard pipeline for training the 120k ImageNet-22k data. The origin images are randomly scaled and cropped to a fixed 224 $\times$ 224 size, and randomly horizontal flipping is used. The total mini-batch size is 256 with 32 images on each GPU and the training epoch is 100. We set the learning rate to 0.1, and it is divided by 10 at 30, 60, and 90 respectively. The total training time is about 25 hours. Finally, we get the pre-trained model with a top-1 accuracy of 69.0 on the \texttt{val} set with respect to the picked categories. The comparison between pre-training and joint-training is present in Table~\ref{tab:pre_training}. The improvement from pre-training is limited, and the overall AP is improved from 25.2 to 26.1. The AP of rare categories only increases by 1.1 points. In contrast, our joint-training pipeline outperforms pre-training by a large margin (3.1 AP). For the rare categories and the common categories, the AP improvements are larger, reaching 8.7 points and 3.5 points respectively. This means that the joint-training is more effective than pre-training when using the additional data for long-tailed object detection. 

\begin{table}
    \centering
    \setlength\tabcolsep{8pt}
    \caption{Comparison between pre-training and joint-training.}
    \begin{tabular}{l | c c c c}
        \toprule
        type & AP & AP\textsubscript{r} & AP\textsubscript{c} & AP\textsubscript{f} \\
        \midrule
        w/o data & 25.2 & 14.6 & 24.0 & 31.2 \\
        pre-training & 26.1 & 15.7 & 25.1 & 31.8\\
        multi-task-training (ours) & \textbf{29.2} & \textbf{24.4} & \textbf{28.6} & \textbf{31.9} \\
        \bottomrule
    \end{tabular}
    \label{tab:pre_training}
\end{table}

\subsubsection{Wild-World Image-level Supervision}

\begin{table}
    \centering
    \setlength\tabcolsep{10pt}
    \caption{Result with wild-world image-level data from Google Images.}
    \begin{tabular}{l | c c c c}
        \toprule
        data & AP & AP\textsubscript{r} & AP\textsubscript{c} & AP\textsubscript{f} \\
        \midrule
        - & 25.2 & 14.6 & 24.0 & 31.2 \\
        Google & 26.8 & 19.9 & 25.7 & 31.0\\
        \bottomrule
    \end{tabular}
    \label{tab:internet_data}
\end{table}

To verify whether CLIS could be applied to a realistic scenario with totally wild-world image-level supervision, we conduct the experiment with the data collected from the Internet. The instance-level dataset $\mathcal{D}_d$ is still set as the LVISv1.0 dataset and the image-level dataset $\mathcal{D}_i$ is collected from the Internet search engine Google Images\footnote{\url{https://www.google.com/imghp?hl=EN.}}. We retrieve images of each category by querying its category names and descriptions provided by the LVISv1.0 dataset. The top 100 images of the 1203 categories are finally picked. By filtering out some noisy data, a total of 96909 images are selected to make up the dataset $\mathcal{D}_i$. As shown in Table~\ref{tab:internet_data}, CLIS could generalize well to the wild-world image-level supervision data and achieves an overall AP of 26.8. What's more, it brings an improvement for the rare categories by 5.3 AP, demonstrating the great potential of leveraging image-level supervision to alleviate the long-tailed object detection problem.

\section{Conclusion}

In this work, we study how to leverage image-level supervision to enhance the long-tailed object detection ability. We propose a novel framework CLIS that tackles the problem in an efficient multi-task learning way. With ingenious design, CLIS learns different tasks collaboratively, sharing the knowledge between image-level supervision and instance-level supervision for better performance of long-tailed object detectors. Extensive experiments on the LVISv1 benchmark demonstrate the effectiveness of CLIS, which establishes a new state-of-the-art. We hope the proposed method could provide new ideas for bringing more data into long-tailed object detection learning.

\section*{Acknowledgments}
This work was supported by the General Program of National Natural Science Foundation of China (NSFC) under Grant 62276189.



 
%

{
  \bibliographystyle{IEEEtran}
  \bibliography{IEEEabrv,egbib}
}

\vfill

\end{document}